\documentclass{article}

\usepackage[preprint, nonatbib]{neurips_2025}

\usepackage[utf8]{inputenc} 
\usepackage[T1]{fontenc}    
\usepackage{hyperref}       
\usepackage{url}            
\usepackage{booktabs}       
\usepackage{amsfonts}       
\usepackage{nicefrac}       
\usepackage{microtype}      
\usepackage{xcolor}         
\usepackage{graphicx}       
\usepackage{amsmath}        
\usepackage{algorithm}      
\usepackage{algorithmic}    
\usepackage{float}          
\usepackage{amssymb}        
\usepackage{nameref}        
\usepackage{wrapfig}        
\usepackage{booktabs}       
\usepackage{subcaption}     

\title{MARché: Fast Masked Autoregressive Image Generation with Cache-Aware Attention}

\author{%
  \textbf{Chaoyi Jiang}$^{*1}$,
  \textbf{Sungwoo Kim}$^{*2}$,
  \textbf{Lei Gao}$^1$, 
  \textbf{Hossein Entezari Zarch}$^1$, \\
  \textbf{Won Woo Ro}$^2$, 
  \textbf{Murali Annavarm}$^1$\\[1em]
  $^1$University of Southern California, $^2$Yonsei University\\
  \texttt{\{chaoyij, leig, entezari, annavara\}@usc.edu}\\
  \texttt{\{sungwoo.kim, wro\}@yonsei.ac.kr}
}

\begin{document}

\maketitle

\begingroup
\renewcommand\thefootnote{*}
\footnotetext{Equal contribution}
\endgroup

\begin{abstract}
Masked autoregressive (MAR) models unify the strengths of masked and autoregressive generation by predicting tokens in a fixed order using bidirectional attention for image generation.
While effective, MAR models suffer from significant computational overhead, as they recompute attention and feed-forward representations for all tokens at every decoding step, despite most tokens remaining semantically stable across steps.
We propose a training-free generation framework \textbf{MARché} to address this inefficiency through two key components: \textit{cache-aware attention} and \textit{selective KV refresh}.
Cache-aware attention partitions tokens into active and cached sets, enabling separate computation paths that allow efficient reuse of previously computed key/value projections without compromising full-context modeling. But a cached token cannot be used indefinitely without recomputation due to the changing contextual information over multiple steps. MARché recognizes this challenge and applies a technique called selective KV refresh. 
Selective KV refresh identifies contextually relevant tokens based on attention scores from newly generated tokens and updates only those tokens that require recomputation, while preserving image generation quality.
MARché significantly reduces redundant computation in MAR without modifying the underlying architecture. 
Empirically, MARché achieves up to 1.7$\times$ speedup with negligible impact on image quality, offering a scalable and broadly applicable solution for efficient masked transformer generation.
\end{abstract}

\section{Introduction}
\label{sec:introduction}

Transformer models have achieved remarkable success in language generation~\cite{vaswani2017attention, brown2020language, raffel2020exploring, touvron2023llama, radford2018improving, radford2019language, ouyang2022training}, spurring their adoption in visual domains such as image synthesis.
Autoregressive image generation models, in particular, generate pixels~\cite{Pixel-1, Pixel-2, Pixel-CNN} or tokens~\cite{parmar2018image, tian2024visual, sun2024autoregressive, DALL-E, Parti, residual-quantization, radford2021learning, ding2021cogview} sequentially by using causal attention to predict one token at a time conditioned on the past. 
Although effective, this strictly sequential process is inherently slow and, more importantly, ill-suited for capturing the two-dimensional spatial structure of images.

To overcome these limitations, masked generative models~\cite{MaskGIT, MUSE, li2023mage} have emerged as a powerful alternative, characterized by parallel token generation and bidirectional attention. 
This leads to improved spatial coherence and significantly faster inference. 
Recently, masked autoregressive (MAR) models~\cite{li2024autoregressive} have demonstrated superior performance by combining the expressive power of autoregressive modeling with the efficiency of masked prediction. 
Unlike prior approaches that rely on vector quantization~\cite{VQ-VAE, VQ-VAE-2, VQGAN}, MAR operates in a continuous token space and employs a diffusion-based loss to model per-token distributions more effectively. 
Leveraging bidirectional attention within this framework, MAR achieves state-of-the-art image generation performance.

Despite its advantages, MAR, similar to masked generative approaches, still incurs substantial computational overhead. 
At each decoding step, the model recomputes attention and feed-forward representations for all tokens, even though only a small subset is selected for generation (also referred to as the unmasked tokens).
Empirically, we observe that a large number of key/value projections are stable across steps, showing minimal variation. This redundancy leads to inefficient computation that limits the scalability of MAR-based generation.

To address this inefficiency, we propose a novel, training-free generation approach \textbf{MARché}, MAR image generation with cache-aware attention.
Our key insight is that token representations in MAR exhibit temporal locality: only a small subset of tokens (those currently being generated or contextually influenced by them) need to be updated, while the rest can safely reuse previously computed attention projections.

MARché consists of two core components: \textit{cache-aware attention} and \textit{selective KV refresh}.
Cache-aware attention is an efficient mechanism that partitions tokens into \textit{active} and \textit{cached} sets, each following a separate computation path.
The active set includes tokens requiring recomputation, while cached tokens reuse their key/value projections from previous steps. The active set consists of tokens that are in need of computing their K and V values, namely all the tokens that have been generated in the previous step and the tokens that are selected for generation in the current step. All the other tokens have their K and V values computed in prior steps and could be reused. However, as we show in our empirical evaluations, reusing cached KV values for all the other tokens is detrimental to the generation quality. Hence, we exploit the structure of MAR to identify several contextually relevant tokens to be placed in the active set, as described in our KV refresh component next. We explicitly decouple the computation for active and cached tokens in both the attention and feed-forward layers, enabling efficient reuse of stable representations while preserving full bidirectional context.

Selective KV refresh complements cache-aware attention by efficiently identifying which tokens require recomputation.
Inspired by previous works that use token-level correlation to manage KV caches~\cite{xiao2023efficient, zhang2023h2o, ge2023model, adnan2024keyformer, chen2024arkvale, tang2024quest}, we analyze attention scores from generating tokens and select only the most contextually relevant ones, defined by their high attention scores, as \textit{refreshing tokens}. These tokens are then included in the active set.
This lightweight mechanism preserves image generation quality while minimizing redundant computation.

Through extensive experiments, we demonstrate that MARché significantly accelerates MAR generation while maintaining high image fidelity. 
Compared to standard MAR,  MARché achieves up to 1.7$\times$ speedup while preserving generation quality. Importantly, MARché requires no architectural changes to the transformer and is fully compatible with existing MAR frameworks, making it broadly applicable in practice.

\section{Related Work}
\label{Related_work}

\paragraph{Image generation with transformers.}Transformer models have been extended to image synthesis by leveraging discrete token representations~\cite{VQ-VAE,VQ-VAE-2,VQGAN},  
which enable language-modeling techniques to be applied to visual data.  
Early approaches~\cite{tian2024visual, sun2024autoregressive, DALL-E, Parti, residual-quantization} adopt autoregressive decoding with causal attention, but struggle to model spatial dependencies.  
To address this, MaskGIT~\cite{MaskGIT} proposes a masked generative model that predicts all tokens in parallel and refines them iteratively, offering improved spatial consistency and faster generation.
Following this, recent masked generative models~\cite{MUSE, li2023mage, weber2024maskbit, lezama2022improved} further improve efficiency and flexibility.
Among them, MAR~\cite{li2024autoregressive} stands out by unifying the strengths of autoregressive and masked modeling: it introduces bidirectional attention into a fixed-order autoregressive framework and eliminates the reliance on vector quantization through a diffusion-based loss.
These optimizations allow MAR to achieve state-of-the-art image generation performance with both higher fidelity and faster inference, and form the baseline approach for this paper.
Other methods, such as VAR~\cite{tian2024visual}, explore hierarchical prediction strategies, but follow a different direction.

\paragraph{Efficient image generation.} Diffusion models have been widely studied from the perspective of efficiency, with strategies including accelerated sampling~\cite{lu2022dpm, lu2022dpm2,liu2023oms}, representation compression~\cite{zhao2024mixdq, zhao2024vidit, yuan2024ditfastattn}, and intermediate result caching~\cite{ma2024deepcache, wimbauer2024cache, agarwal2024approximate}. 
In contrast, efficiency in transformer-based generative models has received less attention despite their high-quality outputs.
For autoregressive transformers, speculative decoding~\cite{jang2024lantern, teng2024accelerating} enables partial parallelism by sampling ahead and verifying predictions, but it is constrained to left-to-right generation and thus orthogonal to our masked setting.

Within masked generative models, several recent approaches have explored efficiency from different perspectives. 
ENAT~\cite{ni2024enat} accelerates decoding by modeling spatial and temporal dependencies between tokens but requires training a dedicated attention module. 
LazyMAR~\cite{yan2025lazymar} speeds up MAR by identifying and skipping redundant token updates based on feature similarity, though it operates externally at the feature level rather than modifying the attention mechanism itself. 
MaGNeTS~\cite{goyal2025masked} improves generation efficiency by dynamically scaling model size during decoding and caching KV pairs, but its caching is tightly integrated with the model's progressive resizing strategy.

KV caching is proposed to improve the efficiency of auto-regressive models~\cite{brown2020language, radford2019language, xiao2023efficient, zhang2023h2o, ge2023model, adnan2024keyformer, chen2024arkvale, tang2024quest, shen2021efficient}. However, the MAR paradigm was not designed to exploit KV caching since the $K$ and $V$ values of a token are repeatedly updated on each step of MAR. MARché makes the key observation that  KV projections of many tokens remain stable across steps, and hence KV caching can in fact be adapted to MAR strategy to improve efficiency. However, the cached KV values have to be occasionally recomputed when the attention scores of cached tokens change. MARché exploits these observations to improve the MAR paradigm without requiring any additional training or architectural modification. 
Crucially, we decouple the attention mechanism into recomputation and reuse phases, enabling selective KV refresh based on token-level relevance. 
By leveraging the structural properties and generation semantics of MAR, MARché achieves both computational efficiency and architectural generality in a lightweight, training-free manner.

\section{Preliminary: Masked Autoregressive Model}
\label{sec:preliminary}

MAR models generate images by iteratively predicting a subset of masked tokens based on previously known ones, following a randomly permuted order. 
This approach generalizes traditional autoregressive decoding by allowing multiple predictions per step while maintaining the autoregressive nature of next-token (or next-set-of-tokens) prediction.

\paragraph{Formulation.}
Let $x = (x_1, x_2, \dots, x_N)$ denote a sequence of image tokens to be generated. 
At the start of inference, all tokens are unknown and initialized as masked, i.e., $x_i^{(0)} = \texttt{[MASK]}$ for all $i$, and the initial set of unknown positions is $M^{(0)} = \{1, 2, \dots, N\}$. 
A random permutation $\pi$ of the index set $\{1, 2, \dots, N\}$ defines the generation order. 
At each generation step $t$, a subset of positions $U^{(t)} \subset M^{(t)}$ is selected for prediction, where $M^{(t)}$ is the current set of masked token positions.

The generation proceeds in two stages: encoding and decoding.  
The encoder $f_{\text{enc}}$ processes only the known tokens, i.e., those at positions $i \notin M^{(t)}$, embedding each with its positional encoding $\text{PE}_i$. 
This yields contextual features for the visible part of the sequence:
\begin{equation}
    h_{\text{enc}} = f_{\text{enc}}(\{x_i^{(t)} + \text{PE}_i \mid i \notin M^{(t)}\}).
\end{equation}

To form the decoder input, we combine $h_{\text{enc}}$ with embeddings $h_{\text{mask}}^{(t)}$ for the masked tokens (i.e., positions in $M^{(t)}$), aligned with their respective positional encodings.
The decoder $f_{\text{dec}}$ attends to the entire sequence and outputs contextual vectors:
\begin{equation}
    z^{(t)} = f_{\text{dec}}(\text{concat}(h_{\text{enc}}, h_{\text{mask}}^{(t)}), \text{PE}).
\end{equation}

For each selected position $i \in U^{(t)}$, a token value is sampled from the conditional distribution modeled by the diffusion sampler, conditioned on the decoder output $z_i^{(t)}$:
\begin{equation}
    x_i^{(t+1)} \sim p_\theta(x_i \mid z_i^{(t)}; \tau),
\end{equation}
where $\tau$ is a temperature parameter that controls the diversity of the generated samples. The predicted values replace the corresponding masked tokens, and the set of unknown positions is updated:
\begin{equation}
    M^{(t+1)} = M^{(t)} \setminus U^{(t)}.
\end{equation}

\vspace{-7pt}
This iterative generation process continues, progressively reducing the number of masked positions, until the entire token sequence is completed, i.e., $M^{(T)} = \emptyset$.

\begin{wrapfigure}{r}{0.3\textwidth}
  \centering
  \includegraphics[width=0.28\textwidth]{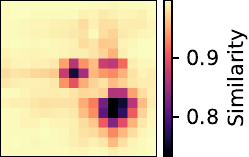}
  \caption{\textbf{Cosine similarity of key projections} between step 4 and 5 (in layer 2) during a single image generation.}
  \vspace{-20pt}
  \label{fig:motivation}
\end{wrapfigure}

\paragraph{Motivation.}
Although MAR models follow an autoregressive decoding paradigm, they use bidirectional attention at each decoding step, which allows \textit{all} tokens, including both generated and yet to be generated ones, to update their contextual representations. 
This design enables global context exchange, but also implies that the features of \textit{all} tokens are recomputed at every step.

However, since only a small subset of tokens is newly generated at each step, the contextual information of the majority of tokens varies minimally between steps.
To quantify this redundancy, we analyze the cosine similarity of key projections between consecutive steps. 
As shown in Figure~\ref{fig:motivation}, a large number of tokens exhibit consistently high similarity, often exceeding 0.95, indicating that their internal representations remain nearly unchanged. 
We observe a similar trend for value projections, reinforcing this observation.

These findings suggest that fully recomputing attention representations for all tokens at every step is inefficient. 
By identifying and reusing stable token representations, we can significantly reduce computational cost.

\begin{figure}[t]
    \centering
    \includegraphics[width=0.9\linewidth]{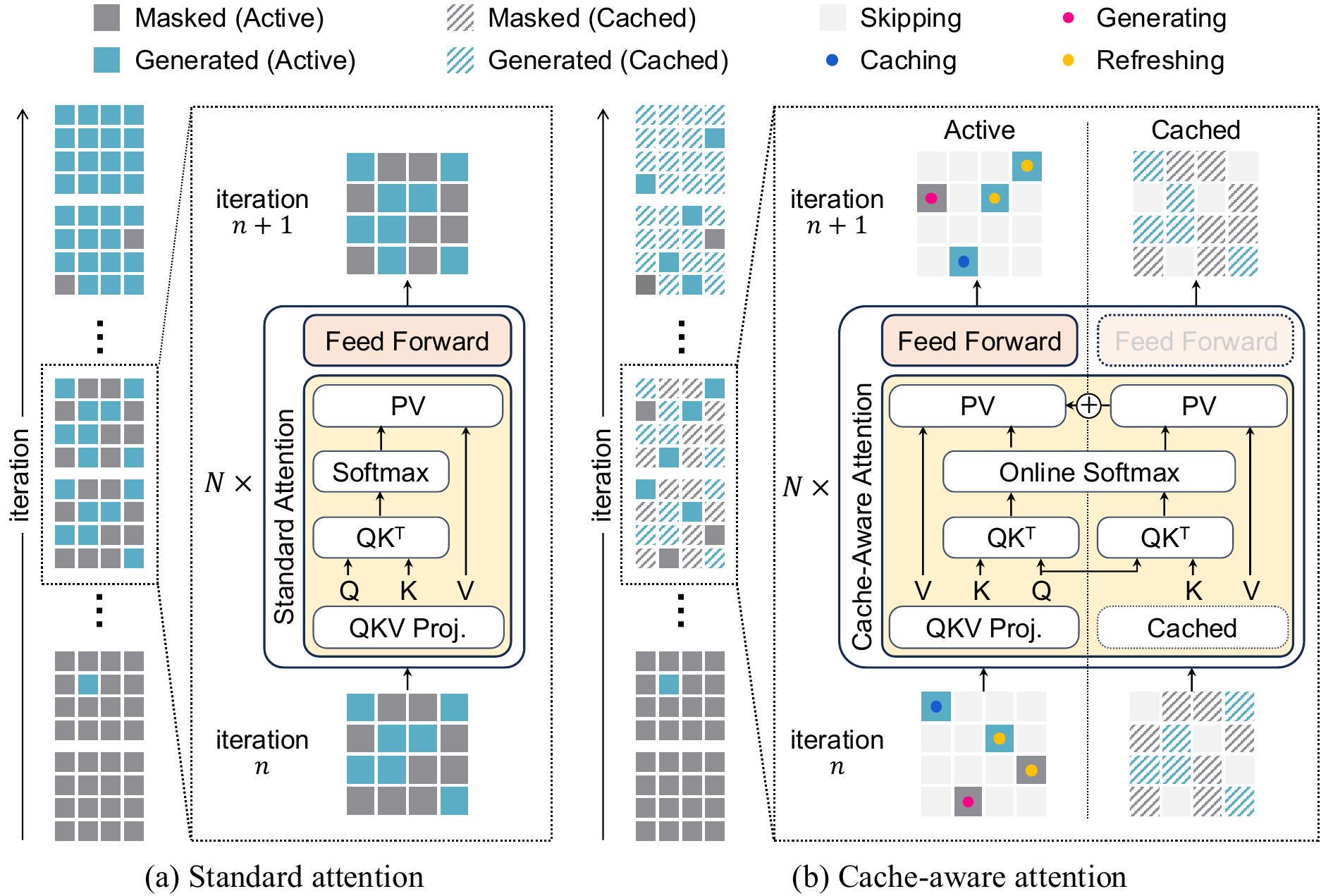}
    \caption{
    \textbf{Comparison of standard attention and cache-aware attention in MAR generation.} 
    \textbf{(a)} In standard attention, all tokens (masked or generated) are processed at every step, leading to redundant computation. 
    \textbf{(b)} In cache-aware attention, only \textit{active tokens} (generating, refreshing, or caching) are recomputed, while \textit{cached tokens} reuse previously computed key/value projections. 
    }
    \label{fig:bidirectional_transformer}
    \vspace{-10pt}
\end{figure}

\section{MARché: Cache-Aware Attention with Selective KV Refresh}
\label{sec:method}

As observed in Section~\ref{sec:preliminary}, many tokens in MAR generation exhibit stable KV projections throughout steps. 
This temporal redundancy suggests an opportunity to reduce the computation by caching and reusing KV projections. 
However, not all tokens are safe to cache, as some exhibit noticeable changes in their contextual representations across steps.

To address this challenge, we introduce \textit{cache-aware attention}, which separates tokens into \textit{active} and \textit{cached} sets, assigning them to distinct computation paths during attention and feed-forward layers (Section~\ref{subsec:attention}).  
Among active tokens, a key subset, \textit{refreshing tokens}, are not being generated but are contextually influenced by generation targets.  
To efficiently identify them, we propose a lightweight attention-guided selection strategy, described in Section~\ref{subsec:refreshment}.



\begin{algorithm}[H]
\caption{Cache-aware Attention}
\label{alg:cache-aware}
\begin{algorithmic}[1]
\REQUIRE Token embeddings $x^{(t)}$, cache $\mathcal{C}^{(t)}$, generation mask $M^{(t)}$
\STATE Identify generating tokens $G^{(t)} \subset M^{(t)}$
\STATE Identify caching tokens $N^{(t)}$ from previous step
\STATE Identify refreshing tokens $R^{(t)}$ based on correlation with $G^{(t)}$
\STATE Set active set $A^{(t)} = G^{(t)} \cup R^{(t)} \cup N^{(t)}$, cached set $C^{(t)} = \{1, \dots, N\} \setminus A^{(t)}$
\STATE Compute $Q, K, V$ for $i \in A^{(t)}$; retrieve cached $K, V$ for $i \in C^{(t)}$
\FORALL{$i \in A^{(t)}$}
    \STATE Compute attention logits over $A^{(t)}$ and $C^{(t)}$, apply safe online softmax:
    \vspace{-5pt}
    \[
    \alpha^{(A)}, \alpha^{(C)}, \ell_i = \texttt{SafeOnlineSoftmax}(q_i, K_A, K_C)
    \]
    \vspace{-10pt}
    \STATE Compute output vector:
    $z_i = \frac{\alpha^{(A)} V_A + \alpha^{(C)} V_C}{\ell_i}$
    \STATE Apply feed-forward: $x_i^{(t+1)} = \text{FFN}(z_i^{(t)})$
\ENDFOR
\STATE Update cache: $\mathcal{C}^{(t+1)}[i] = (K_i, V_i)$ for $i \in A^{(t)}$
\end{algorithmic}
\end{algorithm}

\vspace{-10pt}
\subsection{Cache-aware Attention Mechanism}
\label{subsec:attention}

\paragraph{Token categorization: active vs. cached.}
At each generation step, we categorize tokens into two groups: \textit{active} tokens and \textit{cached} tokens. 
Active tokens are those whose contextual representations must be updated; cached tokens are considered to have very limited changes to their KV projections and can safely reuse their cached key/value (KV) projections.

The active set includes three types of tokens:  
(1) \textit{Generating tokens}, which are masked tokens selected for prediction at the current step;  
(2) \textit{Caching tokens}, which are newly generated tokens from the previous step that have not yet been incorporated into the cache; and  
(3) \textit{Refreshing tokens}, which are not currently being generated but are contextually influenced by the new predictions and thus require updated projections.  
Note that these definitions are used consistently throughout the paper and play a central role in the ablation analysis in Section~\ref{subsec:ablation_study}.

While generating and caching tokens can be identified deterministically, refreshing tokens are selected based on their correlation with generating tokens, as detailed in Section~\ref{subsec:refreshment}.  
All remaining tokens are treated as cached. This categorization forms the foundation of our efficient decoding approach: instead of recomputing all token representations, cache-aware attention assigns computation only to tokens that truly require it.



\paragraph{Cache-aware attention design.}
As illustrated in Figure~\ref{fig:bidirectional_transformer}, cache-aware attention achieves efficiency by splitting the attention operation into two computation paths—one for active tokens and one for cached tokens.  
Active tokens project fresh queries, keys, and values, and compute attention over the union of both active and cached tokens.  
Cached tokens, by contrast, do not generate new projections and are not used as queries, but their stored key/value projections contribute as context during attention.  
This separation allows us to reuse representations where possible, without sacrificing the bidirectional nature of the model.




For each active token, attention is computed in two parts.  
In the first, queries attend to freshly computed keys and values from the active set ($K_A$, $V_A$); in the second, the same queries attend to cached keys and values from previous steps ($K_C$, $V_C$).  
To avoid data movement overhead from concatenation, we compute these two attention scores independently and merge the results using an \textit{safe online softmax} formulation~\cite{FlashAttention-1,FlashAttention-2,FlashAttention-3}.  
With $\alpha^{(A)}, \alpha^{(C)}$, and $\ell_i$ from the online softmax, the output vector is computed as a sum of contributions from active and cached tokens, avoiding the need for key/value concatenation.
This ensures exact equivalence to standard attention while enabling better memory locality and kernel fusion.

After attention, only active tokens are forwarded through the feed-forward network, while cached tokens are skipped entirely in this stage.  
Finally, the key/value projections of active tokens are written back to the cache.  
These design choices of query-side separation, feed-forward skipping, and selective cache updates enable significant runtime savings without modifying the underlying transformer architecture.
The full algorithm is summarized in Algorithm~\ref{alg:cache-aware}.  
We provide a mathematical formulation and proof of equivalence in Appendix~\ref{appendix:equivalence}, and report empirical speedups in Appendix~\ref{appendix:runtime}.

\subsection{Selective KV Cache Refresh}
\label{subsec:refreshment}

\paragraph{Criteria for selecting refreshing tokens.}
As discussed in Figure~\ref{fig:motivation}, although most tokens exhibit stable KV representations across steps, a subset of tokens plays a crucial role in maintaining image generation quality—particularly those influenced by newly generated content. Naturally, newly generated tokens must be recomputed. 
However, tokens that are influenced by the newly unmasked tokens may also need to be refreshed, as their importance can shift dynamically during generation.

To identify such tokens, we draw inspiration from autoregressive language models~\cite{xiao2023efficient, zhang2023h2o, ge2023model, adnan2024keyformer, chen2024arkvale, tang2024quest}, 
where attention scores are commonly used to determine which cached key/value entries can be safely discarded or ignored during inference. 
In contrast, we repurpose attention scores in our setting to do the opposite: to identify which tokens should be actively refreshed. 
By selecting tokens that receive high attention from newly generated tokens, we can target the refresh process toward contextually important tokens.
This strategy enables us to maintain image generation quality with minimal recomputation. 
We validate its effectiveness in Section~\ref{subsec:ablation_study}.


\begin{figure}[t]
    \centering
    \includegraphics[width=0.95\linewidth]{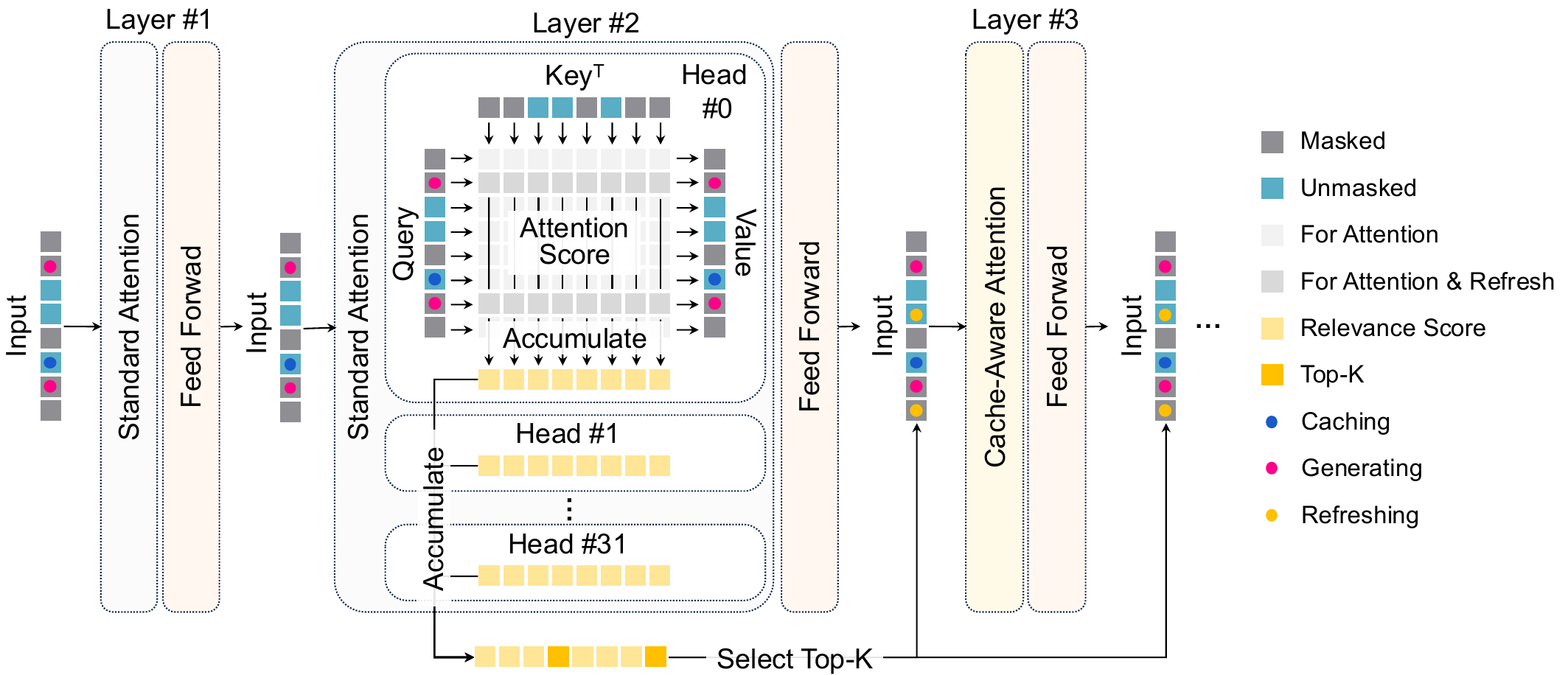}
    \caption{
    \textbf{KV cache refresh via attention score.} 
        Layers 1 and 2 perform standard full attention.
        In Layer 2, attention scores are computed from generating tokens to all others, and aggregated across attention heads to produce global relevance scores. 
        The top-$K$ most contextually relevant tokens are selected based on these scores and marked as refreshing tokens. 
        From Layer 3 onward, cache-aware attention is applied.
    }
    \label{fig:kv_refresh}
\end{figure}

\paragraph{Design of selective KV cache refresh.}
\label{par:selective_design}
Building on our empirical findings, we design a selective KV cache refresh mechanism that dynamically identifies which tokens require recomputation based on their contextual relevance to the generating tokens. 
As illustrated in Figure~\ref{fig:kv_refresh}, we leverage attention scores computed in the second decoder layer (Layer 2). 
Note that the choice of decoder layer for this selection is configurable, allowing for a trade-off between speedup and image quality, which we discuss further in Section~\ref{subsec:ablation_study}.

At each generation step, Layers 1 and 2 perform standard full attention, as the active tokens are not yet finalized. 
During the Layer 2 attention computation, attention scores from generating tokens to all other tokens are computed and aggregated across all attention heads to obtain a global relevance score for each token. 
We then select the top-$K$ tokens with the highest relevance scores as the refreshing tokens. These refreshing tokens, together with the newly generated tokens and the previously cached tokens, form the active token set for subsequent layers. In these layers, cache-aware attention is applied: representations are recomputed only for active tokens, while the KV projections of non-active (cached) tokens are reused without reprocessing. 
This targeted refresh strategy enables the model to preserve contextual consistency while minimizing computational overhead.

\subsection{Implementation Details}
\label{sec:implementation}

\paragraph{Updating active tokens.}

At each generation step, the active token set always includes tokens that are selected for generation in the current step and tokens that were selected for generation in the immediately prior step. We then use the attention scores of the first decode layer to select the top-K tokens to fill a fixed batch size of active tokens. In our current implementation, the batch size is set to 64. Thus, if there are 5 tokens that are selected for generation in the current step, and 9 tokens were selected in the immediately prior step, then we select the remaining 50 tokens as the refresh tokens based on the attention scores.
The number of refreshing tokens is thus adjusted accordingly to fit within this budget. An ablation study on this value is provided in Section~\ref{subsec:ablation_study}.

\paragraph{Refreshing entire KV Cache.}

We refresh the entire KV cache in three specific cases to ensure correctness and stability. 
First, during the initial step, since no KV projections have been generated yet, we apply full attention across all layers. 
Second, in every step, the first and second decoder layers perform full attention as the active tokens are not yet finalized.
Lastly, to mitigate potential value drift caused by repeated caching, we periodically refresh the entire KV cache every three steps, a process we refer to as periodic full refresh.
Ablation results analyzing the impact of full refresh frequency and the choice of refresh layers are provided in Appendix~\ref{appendix:refresh_steps} and Appendix~\ref{appendix:refresh_layers}, respectively.

\section{Evaluation}
\label{sec:evaluation}

We evaluate MARché along three key dimensions: (1) image quality, to ensure that generation fidelity is preserved; (2) inference latency, to assess computational efficiency; and (3) design choices, to analyze how different components of our method impact performance.


\subsection{Experimental Setup}
\label{sec:experimental_setup}

We evaluate our proposed MARché framework on the ImageNet dataset at $256 \times 256$ resolution in the class-conditional setting.
We adopt three model scales from the original MAR implementation~\cite{li2024autoregressive}: MAR-B (Base), MAR-L (Large), and MAR-H (Huge). 
For each model, we construct a corresponding MARché variant (MARché-B, MARché-L, MARché-H) by applying our cache-aware inference strategy without modifying the model architecture or training procedure.

To assess image quality, we report Fréchet Inception Distance (FID)~\cite{heusel2017gans} and Inception Score (IS)~\cite{salimans2016improved}. 
Computational efficiency is measured using the latency metric (in seconds per image). 
Speedup is calculated as the ratio of latency between the original MAR and its corresponding MARché variant at each model scale.

We compare MARché against several strong baselines: MaskGIT~\cite{MaskGIT}, a masked generative transformer; DiT~\cite{DiT}, a diffusion transformer; and LlamaGen~\cite{sun2024autoregressive}, an autoregressive model. 
All experiments are conducted on a single NVIDIA H100 GPU. 
We use a batch size of 128 for main evaluations and 256 for ablation studies. 
We follow the default MAR inference schedule of 64 decoding steps.

\subsection{Main Results}

Table~\ref{tab:main-eval} presents a comprehensive comparison of our MARché models against recent high-performing image generation baselines, including MaskGIT~\cite{MaskGIT}, DiT~\cite{DiT}, and LlamaGen~\cite{sun2024autoregressive}, as well as the original MAR models across three different scales.

Across all model sizes, MARché achieves substantial reductions in inference latency while maintaining competitive generation quality. 
For instance, MARché-B reduces latency from 0.104s to 0.064s per image, achieving a $1.57\times$ speedup over MAR-B, with negligible change in FID (from 2.35 to 2.56) and Inception Score (from 281.1 to 270.3). 
Similar trends are observed for the larger models: MARché-L and MARché-H show $1.68\times$ and $1.72\times$ speedups respectively, while maintaining image quality within acceptable margins. Appendix~\ref{appendix:generated_images} presents qualitative comparisons showing that MARché-H produces images that are visually indistinguishable from those of MAR-H. The FID and IS scores of MARché variants remain close to those of the original MAR models, and are consistently better than those of MaskGIT and LlamaGen.  While DiT achieves slightly better FID than MARché-B, it does so at the cost of significantly higher latency. 

Overall, MARché offers a compelling trade-off between generation speed and quality, outperforming baseline masked and autoregressive models in both inference efficiency and scalability. 
Importantly, these gains are achieved without modifying the original architecture or retraining, demonstrating the practical applicability of our approach.

\begin{table}[t]
\centering
\caption{\textbf{Comparison of MARché and baseline models on ImageNet 256×256.}
MARché achieves up to $1.72\times$ speedup over the original MAR models while maintaining competitive image quality. }
\label{tab:main-eval}
\begin{tabular}{lccccr}
\toprule
\textbf{Method} & \textbf{Latency (s/im) ↓} & \textbf{FID ↓} & \textbf{IS ↑} & \textbf{Param} & \textbf{Speedup ↑} \\
\midrule
MaskGIT~\cite{MaskGIT}                      & 0.440 & 6.18   & 182.1   & 227M  & - \\
DiT-XL/2~\cite{DiT}                         & 0.787 & 2.27   & 278.2   & 675M  & -\\
LlamaGen-XXL~\cite{sun2024autoregressive}   & 0.897 & 3.09   & 253.6   & 1.4B  & - \\
LlamaGen-3B~\cite{sun2024autoregressive}    & 1.011 & 3.05   & 222.3   & 3.1B  & - \\
\toprule
MAR-B~\cite{li2024autoregressive}           & 0.104 & 2.35   & 281.1  & 208M  & 1.00 \\
MARché-B                                    & 0.064 & 2.56   & 270.3  & 208M  & 1.57 \\
\midrule
MAR-L~\cite{li2024autoregressive}           & 0.193 & 1.84   & 296.3  & 479M  & 1.00 \\
MARché-L                                    & 0.115 & 2.16   & 278.6  & 479M  & 1.68 \\
\midrule
MAR-H~\cite{li2024autoregressive}           & 0.336 & 1.62   & 298.6  & 943M  & 1.00 \\
MARché-H                                    & 0.195 & 2.02   & 281.4  & 943M  & 1.72 \\
\bottomrule
\end{tabular}
\end{table}

\subsection{Ablation Study}
\label{subsec:ablation_study}

\paragraph{Ablation on token types in active set construction.}
We validate the design of our active token selection strategy in MARché by comparing four different approaches in Table~\ref{table:selection-strategies}. 
The compared strategies are:
(1) the full MARché method;
(2) MARché without \textit{caching tokens};
(3) MARché without \textit{generating tokens};
(4) random selection of tokens for recomputation.

As shown in Table~\ref{table:selection-strategies}, the full MARché configuration achieves the best performance with an FID of 2.56.
Including caching tokens results in  some performance improvement and more importantly that inclusion costs no additional computational burden. 
In contrast, excluding generating tokens leads to a substantial degradation in quality, confirming their essential role in preserving semantic consistency during decoding.
Finally, using randomly selected tokens for recomputation yields the worst performance, highlighting the importance of guided, attention-based selection.

These results demonstrate that generating tokens are indispensable for accurate generation, while incorporating caching tokens offers additional benefits with minimal computational cost.

\begin{table}[t]
\vspace{-13pt}
\centering
\begin{minipage}[t]{0.48\textwidth}
\centering
\caption{\textbf{Effect of token selection strategies for constructing the active set.}
The results illustrate generation performance with respect to the inclusion of \textit{generating tokens} and \textit{caching tokens}, as defined in Section~\ref{subsec:attention}.}
\label{table:selection-strategies}
\begin{tabular}{lc}
\toprule
\textbf{Strategy} & \textbf{FID ↓} \\
\midrule
MARché              & \textbf{2.56} \\
MARché w/o \textit{caching tokens}                   & 2.70 \\
MARché w/o \textit{generating tokens}                      & 504.82 \\
Random selection                         & 564.61 \\
\bottomrule
\end{tabular}
\end{minipage}
\hfill
\begin{minipage}[t]{0.48\textwidth}
\centering
\caption{\textbf{Effect of refreshing token selection strategy.} Our MARché approach, which selects tokens receiving high attention from generating tokens, outperforms both low-attention and random selection strategies, demonstrating the effectiveness of relevance-based refreshing.}
\label{table:refreshing-ablation}
\begin{tabular}{@{}lc@{}}
\toprule
\textbf{Strategy} & \textbf{FID ↓} \\
\midrule
MARché         & \textbf{2.56} \\
MARché w/ low attention scores                 & 3.80 \\
MARché w/ random selection                        & 3.01 \\
\bottomrule
\end{tabular}
\end{minipage}
\vspace{-12pt}
\end{table}

\paragraph{Correlations of refreshing tokens and attention scores.}
We evaluate the impact of different strategies for selecting \textit{refreshing tokens}. As shown in Table~\ref{table:refreshing-ablation}, we compare three approaches.

The first strategy follows MARché’s approach, selecting refreshing tokens with high attention scores to generating tokens  
The second selects tokens with low attention scores to generating tokens, and the third randomly selects a subset of tokens for refreshing, ignoring contextual relevance.

As the results indicate, selecting tokens based on low attention scores leads to worse performance than random selection, with a clearly higher FID.  
This suggests that refreshing tokens should be chosen based on their strong correlation with generating tokens in order to maintain high image generation quality.

\paragraph{Choosing the decoder layer for refreshing token selection.}

We analyze the impact of selecting different decoder layers for computing attention scores used in refreshing token selection. 
This decision involves a trade-off between computational efficiency and generation quality, as it affects both which tokens are refreshed and where full attention is computed.

\begin{figure}[t]
    \centering
    \begin{minipage}{0.48\linewidth}
        \centering
        \includegraphics[width=\linewidth]{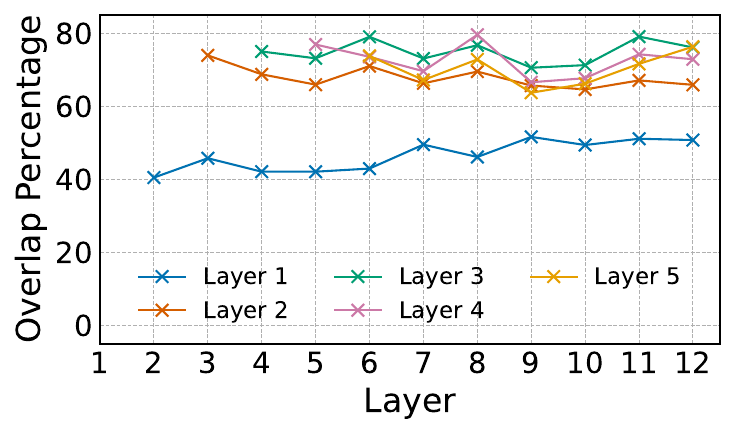}
        \caption{\textbf{Top-$K$ attention overlap across decoder layers.} Deeper layers (3–5) show higher consistency overall, while Layer 1 exhibits much lower overlap (49.3\%).}
        \label{fig:refresh_layer_overlap}
    \end{minipage}
    \hfill
    \begin{minipage}{0.48\linewidth}
        \centering
        \includegraphics[width=\linewidth]{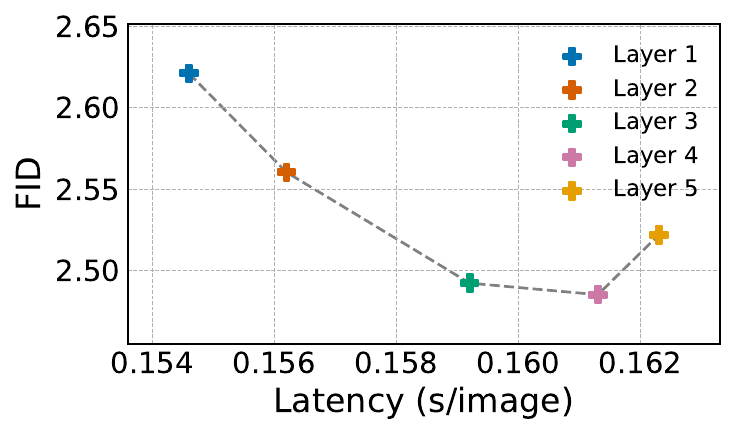}
        \caption{\textbf{Latency–FID trade-off for different refreshing token selection layers.} Deeper layers improves FID with higher latency, while earlier layers decode faster but degrade quality.}
        \label{fig:refresh_layer_FID}
    \end{minipage}
    \vspace{-15pt} 
\end{figure}

As shown in Figure~\ref{fig:refresh_layer_overlap}, deeper layers tend to produce refreshing token selections that align more closely with other layers. 
Specifically, Layer 3 achieves the highest average overlap of 74.4\% with other layers, while Layer 1 shows lower alignment at 49.3\%. 
Layer 2 exhibits a moderate level of agreement at 66.8\%, suggesting it captures context reasonably well while still remaining computationally efficient.

This trade-off is also reflected in generation performance. 
Figure~\ref{fig:refresh_layer_FID} shows that using Layer 1 results in faster decoding (0.154s/image) but yields a relatively high FID of 2.62. 
On the other hand, selecting a deeper layer like Layer 4 improves FID to 2.49 but but with increased latency (0.1613s/image). 
Layer 2 provides a balanced compromise, attaining an FID of 2.56 with moderate latency (0.1593s/image).

In our experiments, we adopt Layer 2 as a default, as it provides a favorable trade-off between speed and quality while maintaining reasonable alignment with other layers’ token selection.

\begin{wrapfigure}{r}{0.43\textwidth}
  \centering
  \vspace{-10pt} 
  \includegraphics[width=\linewidth]{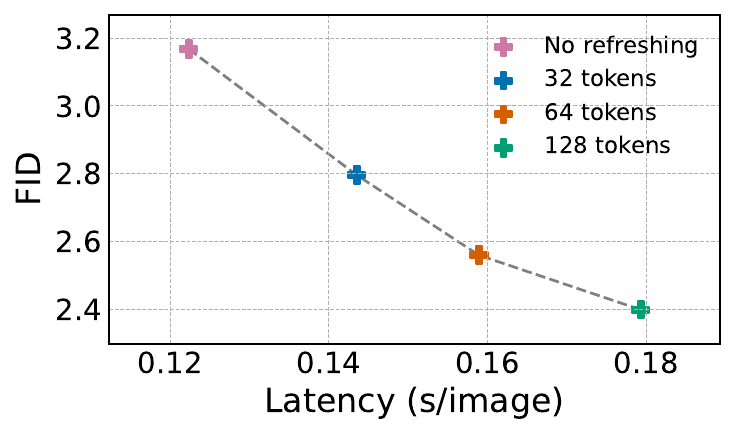}
  \caption{\textbf{Trade-off between latency and FID as a function of the number of refreshing tokens.}
Using more refreshing tokens improves FID but increases latency.}
  \label{fig:refreshing_token}
  \vspace{-15pt} 
  
\end{wrapfigure}

\paragraph{Effectiveness of the number of refreshing tokens.}

We analyze how the number of refreshing tokens affects the trade-off between generation quality and latency. 
In MARché, the number of refreshing tokens is determined by the total number of active tokens minus the generating and caching tokens. 
By varying the number of active tokens, we can control how many tokens are refreshed at each step.

As shown in Figure~\ref{fig:refreshing_token}, increasing the number of refreshing tokens improves image quality, lowering FID, but also increases latency due to additional computation. 
Among the tested settings, using 64 refreshing tokens provides a good balance, achieving low FID with moderate latency.

We also evaluate a configuration with no refreshing tokens, which yields the fastest inference (0.12s/image), but results in a notable drop in image quality, with FID rising to 3.17.
This highlights the necessity of refreshing contextually relevant tokens for maintaining generation fidelity.

\section{Conclusion}
\label{sec:conclusion}

We presented MARché, a cache-aware decoding framework for masked autoregressive image generation that significantly reduces inference latency while maintaining high image quality. 
By selectively refreshing only contextually relevant tokens based on attention scores, MARché avoids redundant computation and achieves up to $1.7\times$ speedup across model scales without modifying the original architecture.
Our approach highlights the potential of combining autoregressive modeling with efficient token reuse strategies in high-resolution image generation. 
While our method focuses on masked autoregressive models, the core ideas behind cache-aware attention and selective refresh may extend to other generative settings, including multi-modal generation.



\medskip

\newpage
\appendix

\section{Equivalence of Cache-Aware Attention to Standard Attention}
\label{appendix:equivalence}

We show that our cache-aware attention mechanism is mathematically equivalent to standard full attention, assuming no approximation in cache usage or token selection. This guarantees that our method maintains the expressivity of full bidirectional attention despite operating in a more efficient, partitioned manner.

\subsection{Standard Attention Formulation}

Let $q_i \in \mathbb{R}^d$ be a query corresponding to token $i$, and $K = [k_1, \dots, k_N]^\top \in \mathbb{R}^{N \times d}$, $V = [v_1, \dots, v_N]^\top \in \mathbb{R}^{N \times d}$ be the full sets of key and value vectors. The standard scaled dot-product attention computes:

\begin{equation}
    \text{Attn}(q_i, K, V) = \sum_{j=1}^{N} \frac{\exp\left( \frac{q_i \cdot k_j}{\sqrt{d}} \right)}{\sum_{\ell=1}^{N} \exp\left( \frac{q_i \cdot k_\ell}{\sqrt{d}} \right)} v_j.
\end{equation}

\subsection{Cache-aware attention formulation}

In cache-aware attention, we partition the key-value set into two disjoint subsets:
\begin{itemize}
    \item $A$: the active token set, with freshly computed $K_A$, $V_A$,
    \item $C$: the cached token set, with previously computed $K_C$, $V_C$,
\end{itemize}
such that $A \cup C = \{1, \dots, N\}$ and $A \cap C = \emptyset$.

Rather than computing attention over the full $K$, $V$ simultaneously, we compute the attention output incrementally using the online softmax trick as follows:

\begin{align}
    s^{(A)} &= \frac{q_i K_A^\top}{\sqrt{d}}, \quad s^{(C)} = \frac{q_i K_C^\top}{\sqrt{d}}, \\
    m_i &= \max(\max s^{(A)}, \max s^{(C)}), \\
    \alpha^{(A)} &= \exp(s^{(A)} - m_i), \quad \alpha^{(C)} = \exp(s^{(C)} - m_i), \\
    \ell_i &= \sum \alpha^{(A)} + \sum \alpha^{(C)}, \\
    z_i &= \frac{\alpha^{(A)} V_A + \alpha^{(C)} V_C}{\ell_i}.
\end{align}

\subsection{Equivalence Proof}

Observe that standard attention can also be rewritten in terms of a shared max $m_i$:

\begin{equation}
    \text{Attn}(q_i) = \frac{ \sum_{j=1}^{N} \exp\left( \frac{q_i \cdot k_j}{\sqrt{d}} - m_i \right) v_j }{ \sum_{j=1}^{N} \exp\left( \frac{q_i \cdot k_j}{\sqrt{d}} - m_i \right) }.
\end{equation}

Since $A$ and $C$ partition $\{1, \dots, N\}$, the sums over $j \in A \cup C$ in cache-aware attention cover exactly the same elements as the full attention formulation. Therefore,

\[
z_i = \frac{ \sum_{j \in A \cup C} \exp\left( \frac{q_i \cdot k_j}{\sqrt{d}} - m_i \right) v_j }{ \sum_{j \in A \cup C} \exp\left( \frac{q_i \cdot k_j}{\sqrt{d}} - m_i \right) } = \text{Attn}(q_i, K, V).
\]

\subsection{Conclusion}

Thus, the cache-aware attention formulation yields exactly the same output as standard attention, up to floating point precision. The separation into active and cached components, combined with the use of an online softmax normalization, offers significant computational advantages without sacrificing model fidelity.

\section{Runtime Efficiency of Cache-Aware Attention Implementation}
\label{appendix:runtime}

\begin{figure}[H]
  \centering
  \includegraphics[width=0.9\linewidth]{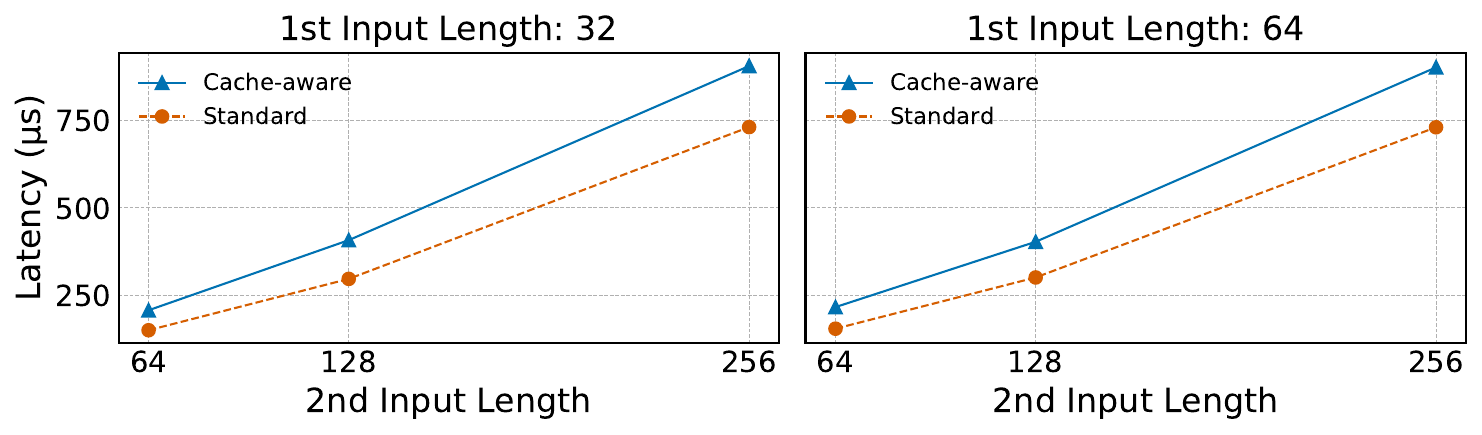}
  \caption{\textbf{Runtime comparison of standard attention vs.\ cache-aware attention implementation.} 
Both methods process the same total number of attention operations over two input segments of different lengths. 
\textit{Standard attention} concatenates the inputs and processes them jointly, while \textit{cache-aware attention} handles them separately. 
}
  \label{fig:runtime_cache_aware}
\end{figure}

In the cache-aware attention operation, the inputs include the query vector $q_i$, active keys and values $K_A$ and $V_A$, and cached keys and values $K_C$ and $V_C$. A vanilla implementation leverages a standard attention kernel by first concatenating $K_A$ with $K_C$, and $V_A$ with $V_C$, to form unified key and value projections. These are then passed to the standard attention kernel for computation. However, we demonstrate that this approach is suboptimal. Assuming all inputs $q_i$, $K_A$, $V_A$, $K_C$, and $V_C$ are preloaded and available, we compare the latency of this baseline with our optimized cache-aware attention kernel. 


Figure~\ref{fig:runtime_cache_aware} reports kernel-level latency across varying input sizes. 
Despite involving the same total number of attention operations, our cache-aware attention kernel consistently achieves lower latency than the vanilla concatenation-based implementation.
For instance, when the two input lengths are 32 and 256, cache-aware attention reduces latency from 904.2$\mu$s to 730.3$\mu$s, yielding a 19.2\% improvement. 
Across all tested configurations, we observe latency reductions ranging from 16.2\% to 27.3\%, with larger input asymmetries showing greater benefit.

These improvements are attributed to reduced memory movement (by avoiding key/value concatenation), improved memory locality, and enhanced kernel fusion. 

\newpage
\section{Effect of Refresh Frequency across Steps}
\label{appendix:refresh_steps}

\begin{wrapfigure}{r}{0.48\textwidth}
  \vspace{-15pt}
  \centering
  \includegraphics[width=\linewidth]{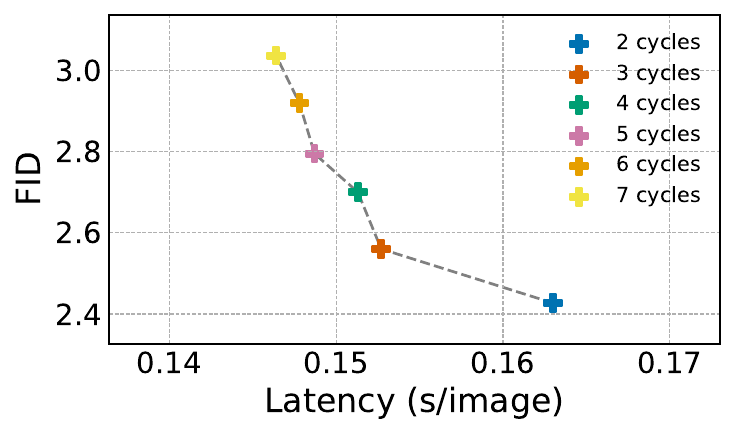}
  \caption{\textbf{Trade-off between refresh frequency and generation performance.} 
    Shorter refresh cycles (e.g., every 2–3 steps) improve FID at the cost of increased latency, while longer cycles reduce latency but degrade image quality. 
    A 3-step refresh strikes a good balance, achieving low FID with faster decoding.}
  \label{fig:refresh_period}
  \vspace{-10pt}
\end{wrapfigure}

By default, MARché performs a full refresh of the KV cache every 3 decoding steps. 
We investigate how the frequency of these periodic refreshes affects generation quality and inference speed. 
Specifically, we vary the refresh period (i.e., how often full refresh is applied during decoding) and evaluate the trade-off between FID and latency.

As shown in Figure~\ref{fig:refresh_period}, shorter refresh cycles (e.g., every 2 steps) yield better image quality with lower FID scores, but incur higher latency due to more frequent computation. 
In contrast, longer cycles (e.g., every 6–7 steps) reduce latency but significantly degrade generation quality. 
Notably, a 3-step refresh period offers a favorable trade-off: it achieves competitive FID while being much faster than 2-step refresh. 
Based on this, we adopt a 3-step refresh as our default configuration.

These results highlight the importance of carefully tuning refresh frequency. 
Overly frequent refresh leads to redundant computation, while infrequent refresh compromises semantic consistency in generation.

\section{Effect of Refresh Location across Layers}
\label{appendix:refresh_layers}

\label{appendix:refreshing}
\begin{table}[H]
\centering
\caption{\textbf{Ablation on full refresh placement across layers.}
We compare four layer-wise full refresh strategies over 12 decoding layers. 
Refreshing only Layer~1 yields the fastest inference, while refreshing Layers~1–6 slightly improves quality (lowest FID and highest IS). Refreshing deeper or non-consecutive layers degrades both quality and speed.}
\label{tab:refresh_layer-wise}
\begin{tabular}{lccc}
\toprule
\textbf{Refresh Strategy} & \textbf{Latency (s/im) ↓} & \textbf{FID ↓} & \textbf{IS ↑} \\
\midrule
Full at layer 1 only              & \textbf{0.155}   & 2.62   & 266.3 \\
Full for layers 1–6              & 0.164   & \textbf{2.49}   & \textbf{275.1} \\
Full for layers 7–12             & 0.177   & 25.35  & 98.50 \\
Full on even layers              & 0.185   & 4.65   & 212.3 \\
\bottomrule
\end{tabular}
\end{table}

We conduct an ablation study to examine how the placement of full refresh across decoder layers affects generation quality and inference speed in MARché. 
Table~\ref{tab:refresh_layer-wise} compares four strategies in which full refresh is applied to different subsets of layers over 12 decoding layers.
Note that Layer 1 always performs full refresh across all strategies, as it is used to select refreshing tokens.

Applying full refresh only at Layer 1 yields the fastest inference (0.158 s/im) while maintaining strong image quality (FID 2.56, IS 270.3). 
Refreshing the early layers (Layers 1–6) slightly improves quality (FID 2.49, IS 275.1) at a small latency cost (0.160 s/im). 
In contrast, refreshing the deeper layers (Layers 7–12) leads to significant degradation in FID (25.35) and IS (98.5), along with increased latency. 
A similar trend is observed when refreshing only the even-numbered layers.

These results suggest that, when full refresh is applied, it is most effective to perform it in the early layers and in consecutive order, rather than scattered or at later layers.

\newpage
\section{Qualitative Comparison on Generated Images}
\label{appendix:generated_images}

\begin{figure}[H]
  \centering
  \includegraphics[width=1\linewidth]{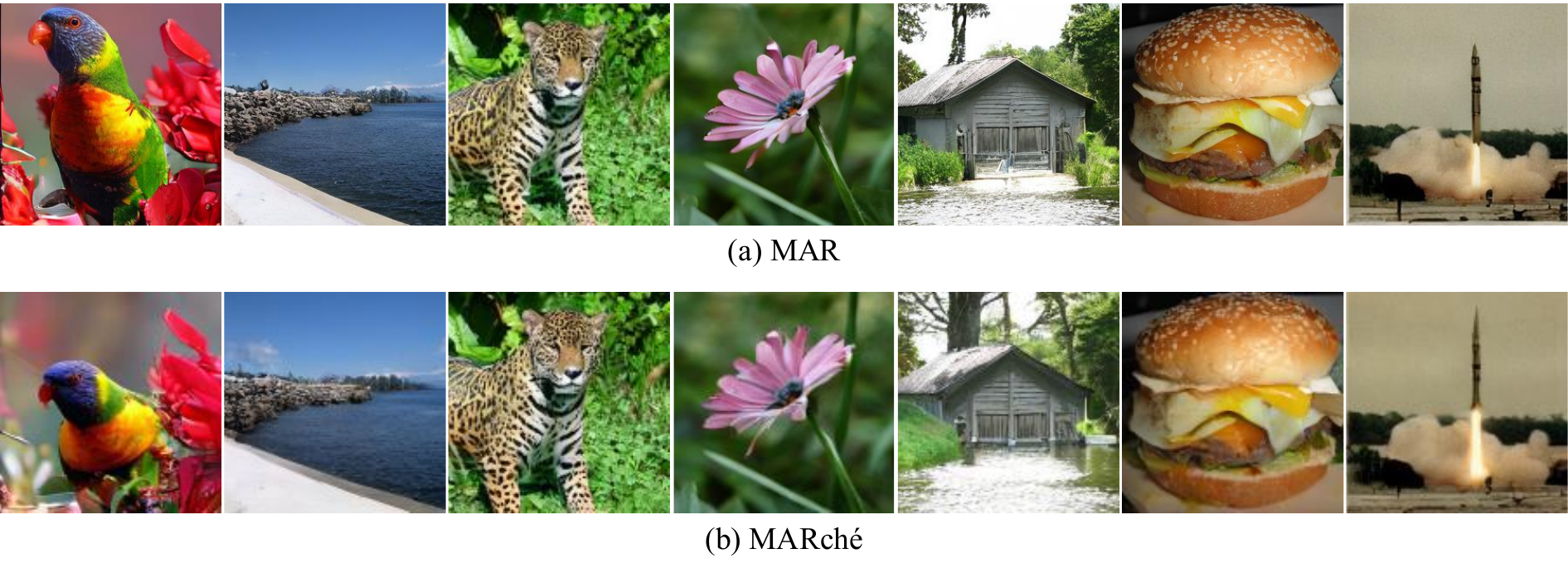}
  \caption{\textbf{Qualitative comparison of generated images from MAR-H (top) and MARché-H (bottom).} MARché-H achieves similar visual quality while running 1.7$\times$ faster.}
  \label{fig:images}
\end{figure}

We conduct a qualitative comparison between standard MAR and our proposed MARché, using the MAR-H and MARché-H, respectively.
As shown in Figure~\ref{fig:images}, despite MARché-H achieving a significant 1.72$\times$ speedup over MAR-H, the visual quality of the generated images remains comparable.
The outputs from MARché-H are virtually indistinguishable from those of MAR, demonstrating that our cache-aware generation approach preserves image fidelity while substantially improving efficiency.
This highlights MARché’s practical value in scenarios requiring both high-quality generation and fast inference.

\section{Stability of Key/Value Projections across Layers and Steps}
\label{appendix:similarity}

\begin{figure}[H]
    \vspace{-20pt}
    \centering
    \includegraphics[width=0.95\linewidth]{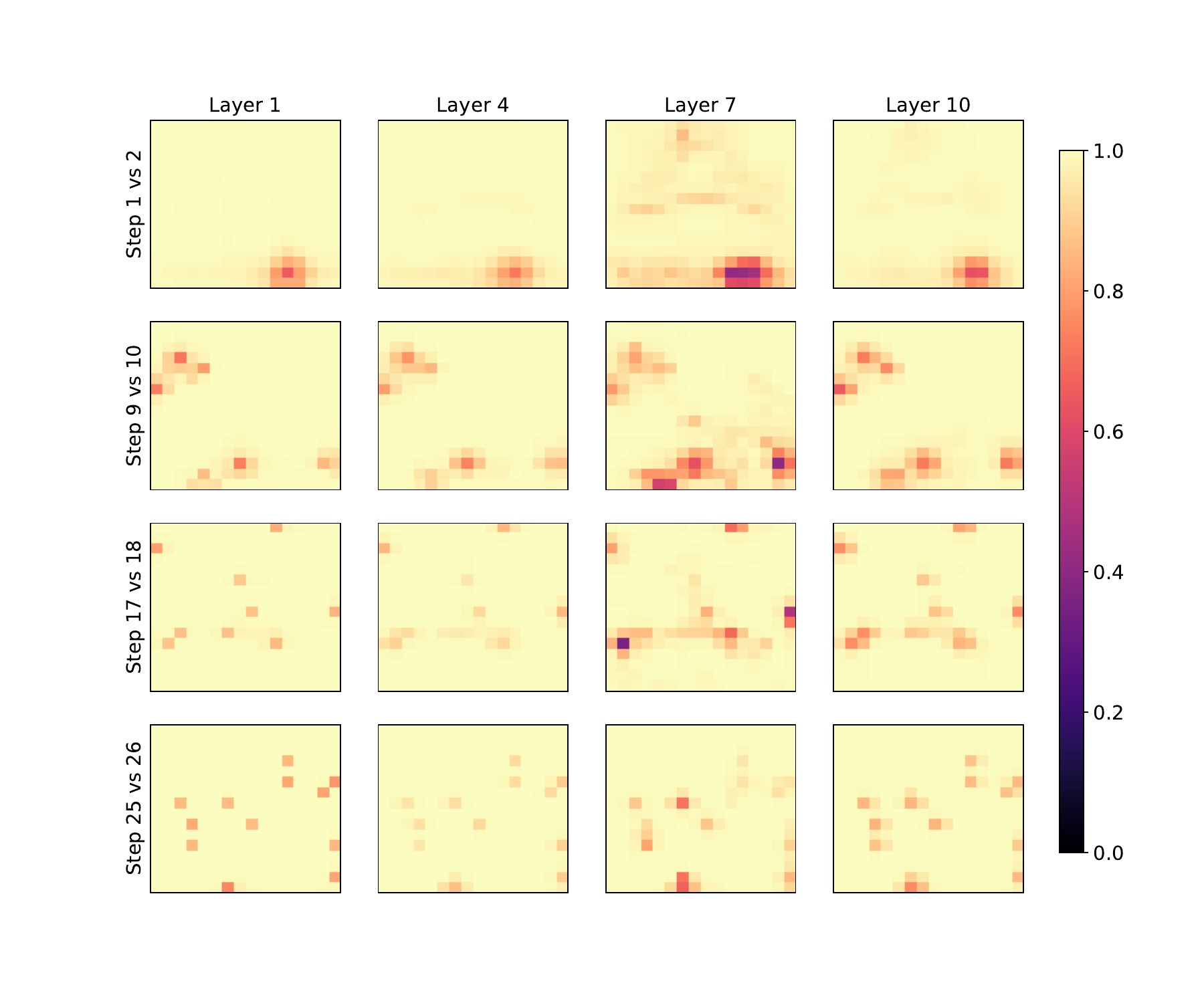}
    \vspace{-30pt} 
    \caption{\textbf{Cosine similarity of key projections across decoding steps and layers.} Most tokens maintain high similarity across steps, especially in lower layers, indicating that their key projections change very little during decoding.}
    \label{fig:k_similarity}
\end{figure}

\begin{figure}[H]
    \vspace{-20pt}  
    \centering
    \includegraphics[width=0.95\linewidth]{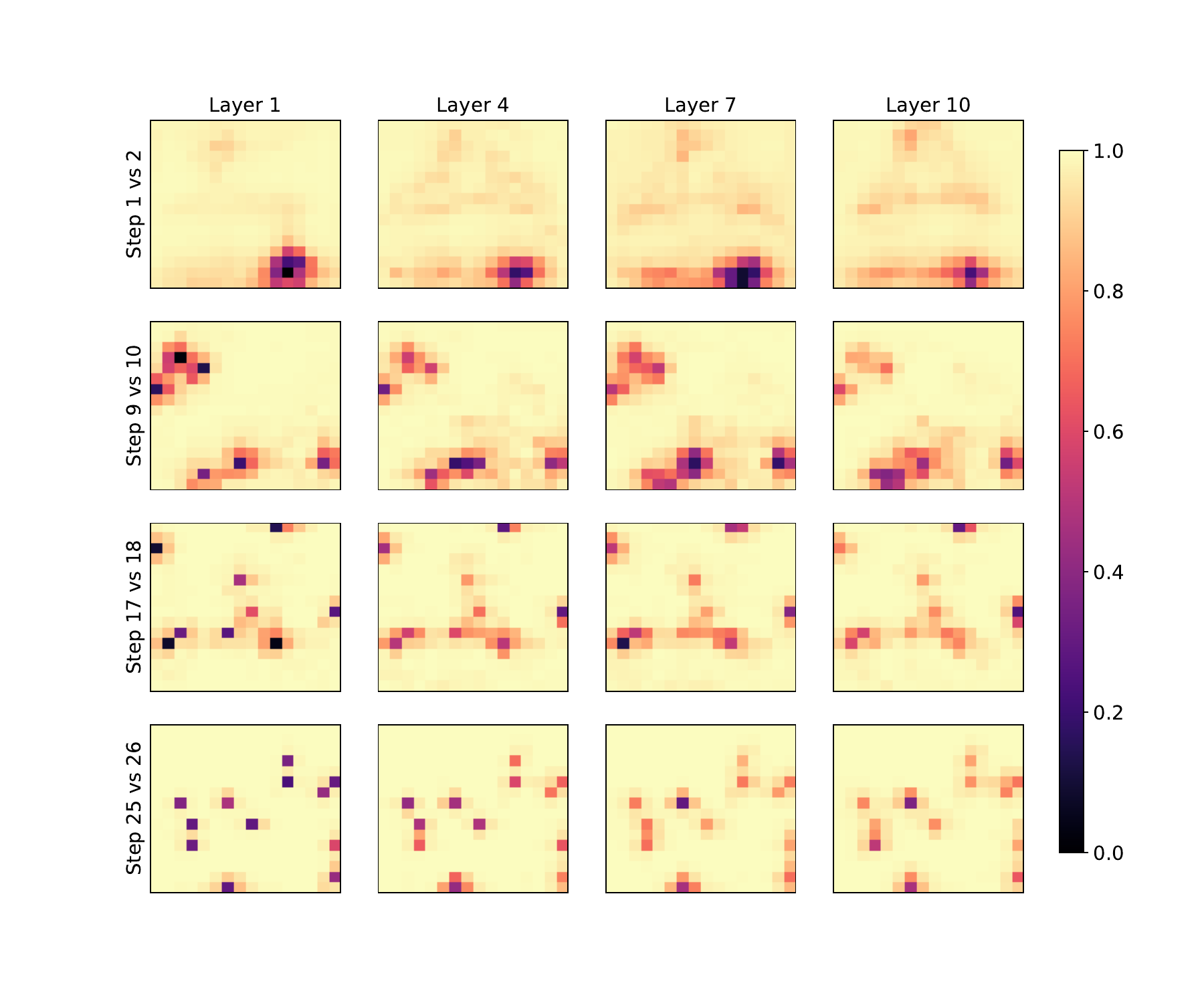}
    \vspace{-30pt} 
    \caption{\textbf{Cosine similarity of value projections across decoding steps and layers.} Value projections are slightly more dynamic than key projections, particularly in deeper layers, but remain stable for the majority of tokens.}
    \label{fig:v_similarity}
\end{figure}

To further understand the temporal redundancy of token representations in MARché, we visualize the cosine similarity of key and value projections between decoding steps across different decoding layers. 
Figures~\ref{fig:k_similarity} and~\ref{fig:v_similarity} show pairwise similarities for selected steps and layers.

\paragraph{Key projections.} 
As shown in Figure~\ref{fig:k_similarity}, the key projections remain highly stable throughout generation, especially in shallow layers (e.g., Layer 1 and Layer 4). 
Even in deeper layers, most tokens exhibit similarities above 0.9 between steps, suggesting minimal variation in their contextual embeddings. 
This supports our motivation that many tokens do not require recomputation at each step.

\paragraph{Value projections.} 
Figure~\ref{fig:v_similarity} reveals slightly more variability in the value projections, particularly in deeper layers and at later steps. 
Nonetheless, the majority of tokens still exhibit high similarity across steps, reinforcing the notion that value projections are also largely stable during decoding.

These findings confirm that both key and value representations show strong temporal locality. This justifies the design of cache-aware attention and selective KV refresh, as only a small subset of tokens truly require recomputation during generation.

\section{Layer-wise Consistency of Refreshing Token Selection Over Step}
\label{appendix:top_k}

\begin{figure}[H]
    \centering
    \includegraphics[width=1\linewidth]{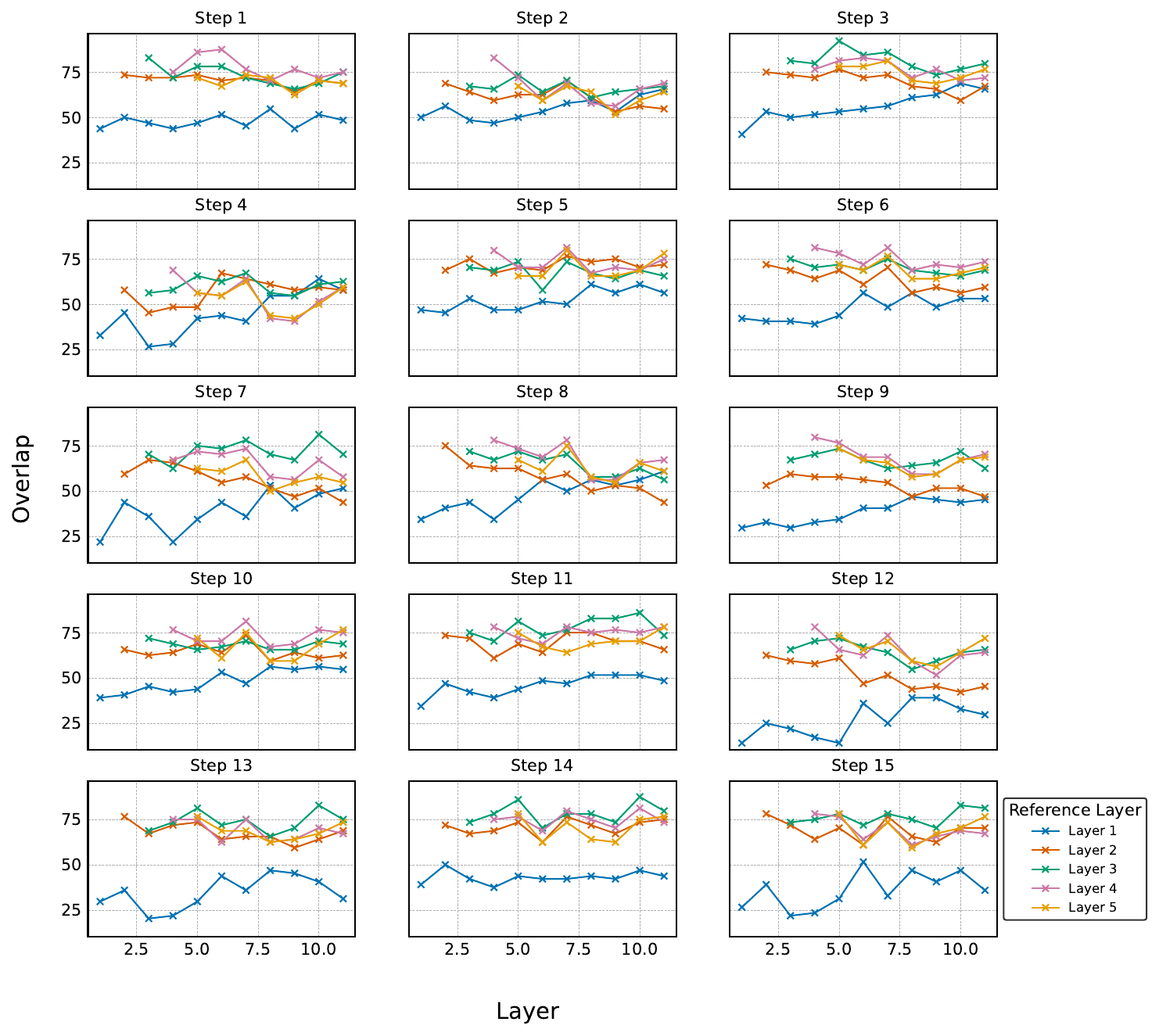}
    \caption{\textbf{Layer-wise top-$K$ attention overlap across decoding steps.} Deeper layers show higher and more consistent alignment.}
  \label{fig:appendix_layerwise_overlap}
\end{figure}

To further analyze the stability of refreshing token selections across layers, we visualize the overlap of top-K attended tokens across decoder layers at each generation step (Steps 1–15), as shown in Figure~\ref{fig:appendix_layerwise_overlap}.
This complements the aggregated results in Figure~\ref{fig:refresh_layer_overlap} by revealing temporal dynamics of layer-wise agreement.

Each subplot in Figure~\ref{fig:appendix_layerwise_overlap} corresponds to a generation step and plots the overlap percentage of different decoder layers with a given reference layer.
Across all steps, we observe that deeper layers (Layers 3–5) consistently exhibit higher mutual alignment compared to shallower ones.
Notably, Layer 1 shows significantly lower agreement with other layers, often falling below 50\% overlap.
In contrast, Layer 3 maintains consistently strong overlap with other deep layers, reaffirming its role as a stable and contextually rich candidate for refreshing token selection.

Layer 2 emerges as a strong middle ground: it achieves higher consistency than Layer 1 while avoiding the computational burden of deeper layers like Layer 4 or 5.
Its overlap trends remain reasonably stable across steps, frequently aligning above 60\% with deeper layers, which suggests that it captures contextual signals effectively without incurring excessive computation.
This observation supports our choice of Layer 2 as the default configuration, offering a practical trade-off between stability and efficiency for refreshing token selection.

\end{document}